\documentclass{article}
\usepackage{arxiv}

\usepackage[utf8]{inputenc} 
\usepackage[T1]{fontenc}    
\usepackage{hyperref}       
\usepackage{url}            
\usepackage{booktabs}       
\usepackage{amsfonts}       
\usepackage{nicefrac}       
\usepackage{microtype}      
\usepackage{lipsum}
\usepackage{graphicx}
\usepackage{threeparttable}
\usepackage{siunitx} 
\usepackage{bm}
\usepackage{subcaption}
\usepackage{multirow}
\usepackage[export]{adjustbox}

\title{Object Manipulation of the Variable Topology Truss system}

\author{
 Andrew Jang-Ho Bae \\
  Department of Mechanical Engineering\\
  University of Nevada, Las Vegas\\
  Las Vegas, NV 89154 \\
  \texttt{andrew.bae@unlv.edu} \\
   \And
 Myeongjin Choi \\
  Advanced Robotics Research Center\\
  Research Institute of AI Robotics\\
  Korea Institute of Machinery and Materials\\
  Daejeon 34103 \\
  \texttt{aud0109@kimm.re.kr} \\
  \And
 Haorui Li \\
  RealMan Robotics Co., Ltd.\\
  Beijing, 100043\\
  \texttt{lihaorui0331@gmail.com} \\
  \And
 Mark Yim \\
  Department of Mechanical Engineering \\ 
  and Applied Mechanics \\
  University of Pennsylvania\\
  Philadelphia, PA 19104 \\
  \texttt{yim@seas.upenn.edu} \\
  \And
 TaeWon~Seo \\
  Department of Mechanical Engineering\\
  Hanyang University\\
  Seoul, 04763 \\
  \texttt{taewonseo@hanyang.ac.kr} \\
}

\begin{document}
\maketitle
\begin{abstract}
This paper presents an object manipulation strategy for the Variable Topology Truss (VTT) system, a truss robot that comprises  actuated truss members  connected by passive spherical joints.
Although truss robots were originally proposed as rapidly deployable manipulators,  manipulation strategy has not been studied thoroughly.
To enable manipulation, we introduce a hybrid control framework that regulates position and force concurrently without explicit decoupling.
At the actuator level, each member  employs a sensor-based force feedback controller to generate the desired axial forces despite high actuator friction.
At the task level, the forces applied at the end-effector nodes are produced by computing the required member forces using a static model of the VTT.
We evaluate force-tracking performance through experiments on both a single member module and the full VTT system.
Finally, we demonstrate object manipulation using two representative configurations and quantitatively assess combined position and force tracking performance. 
Experimental results confirm that the proposed approach enables consistent and reliable object manipulation with the VTT system.
\end{abstract}


\section{Introduction}
Truss robot systems have been studied for a considerable to exploit their versatility and flexibility.
These systems comprise prismatic joints as truss member modules and are connected together by passive spherical joints.
The first truss robot system  applications were as a space-deployable manipulator system \cite{miura1985variable} including  a hardware prototype \cite{hughes1991trussarm}.
Later studies focused on the use of truss robot systems for mobile exploration  \cite{hamlin1997tetrobot, lee2002dynamic, curtis2007tetrahedral}.

Although truss robot systems have been proposed for use as deployable manipulators, their manipulation capabilities have not yet been well demonstrated.
Several early studies on truss robot systems presented kinematic analysis and trajectory planning algorithms for manipulation \cite{naccarato1991inverse, chen1993adaptive}.
Based on these studies, Chirikjian and Burdick performed a 2-dimensional truss robot maneuver and showed obstacle avoidance and manipulation \cite{chirikjian1994hyper}.
Recent studies also focused on trajectory planning for truss robot systems \cite{zhao2016inverse, shen2017space}.
Usevitch et al.~proposed a truss-like robotic system made of soft materials and proposed a manipulation method that involves grabbing objects between the soft edges of the structure \cite{usevitch2020untethered, hammond2021grasp}.
Although the proposed untethered soft robot is not a typical rigid truss robot system, the study demonstrated the potential for manipulation of truss-like robots.

Force and position control, especially in task space, is crucial for performing object manipulation.
In truss-robot systems, task-space control refers to regulating the position and/or force of a target node.
Although task space force control is clearly beneficial for object manipulation with truss robot systems, no studies have specifically focused on its implementation.
Several previous studies focused on the dynamic modeling and analyzed manipulator-level force of truss robot systems with simulation results \cite{furuya1995dynamics, boutin1999dynamics, rost2012design}.
Cubero and Billingsley investigated force control for a truss robot system, but their experiments were limited to a single pneumatic actuator \cite{cubero1997force}.

\begin{figure}[t]
\centering

\begin{minipage}[t]{0.31\textwidth}\vspace{0pt}\centering
  \begin{subfigure}[t]{\linewidth}
    \includegraphics[width=\linewidth]{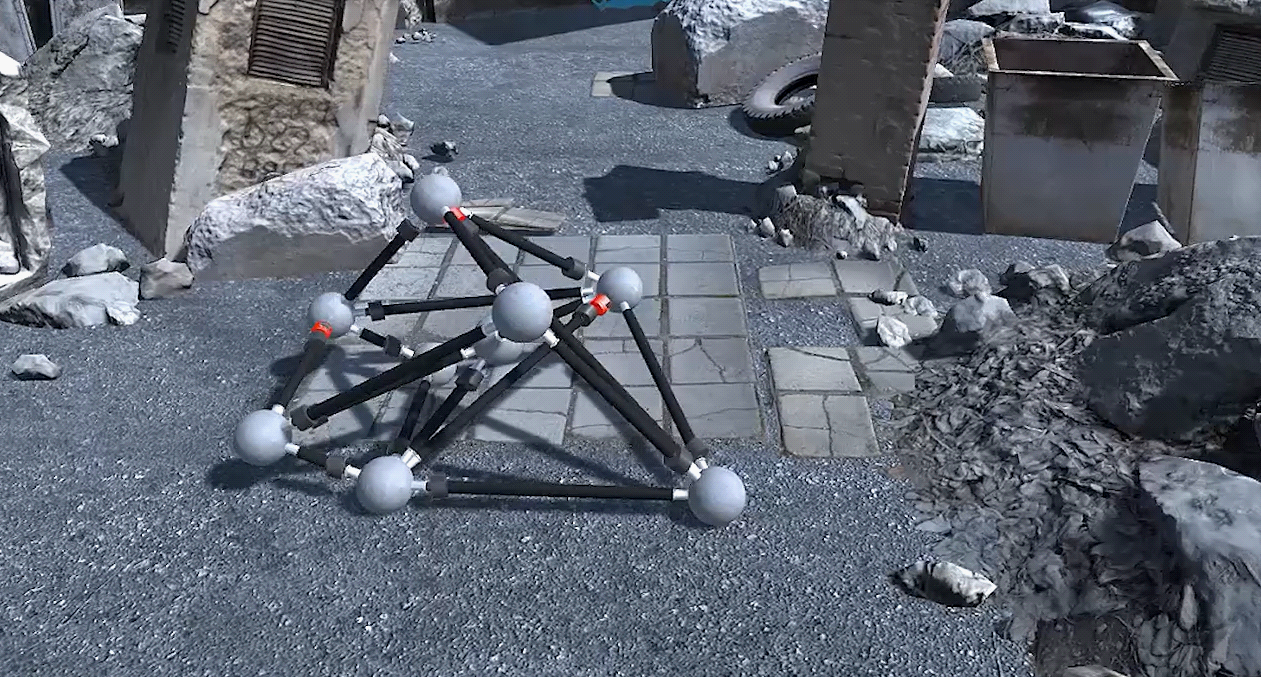}
    \caption{}
    \label{Fig_1_a}
  \end{subfigure}\vspace{3pt}
  \begin{subfigure}[t]{\linewidth}
    \includegraphics[width=\linewidth]{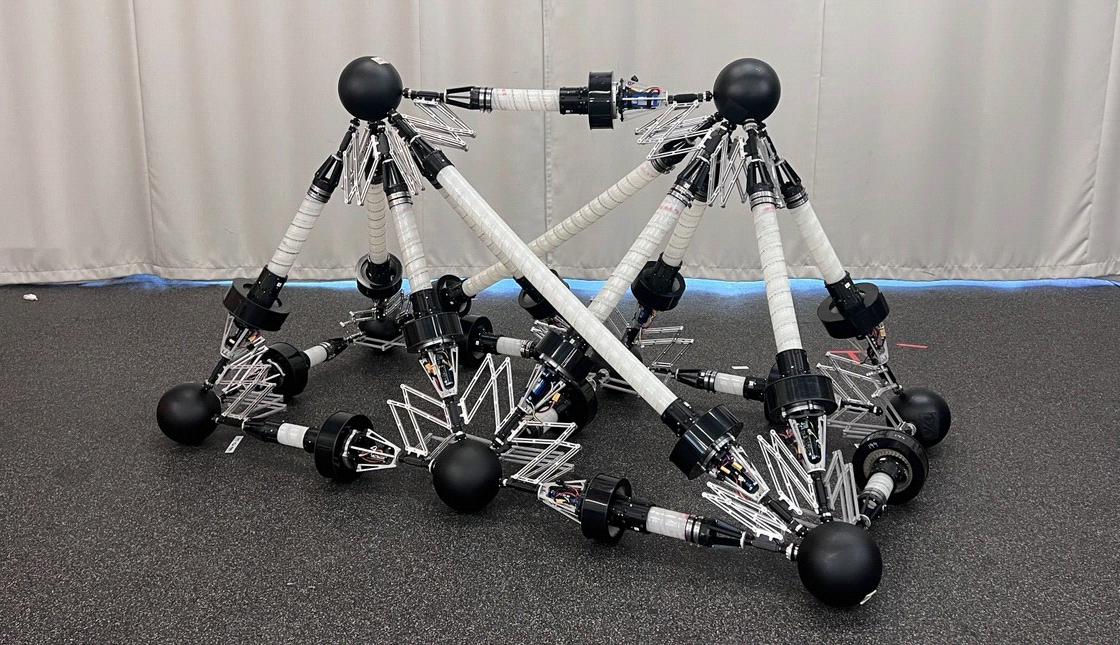}
    \caption{}
    \label{Fig_1_b}
  \end{subfigure}
\end{minipage}
\hfill
\begin{minipage}[t]{0.259\textwidth}\vspace{0pt}\centering
  \begin{subfigure}[t]{\linewidth}
    \includegraphics[width=\linewidth]{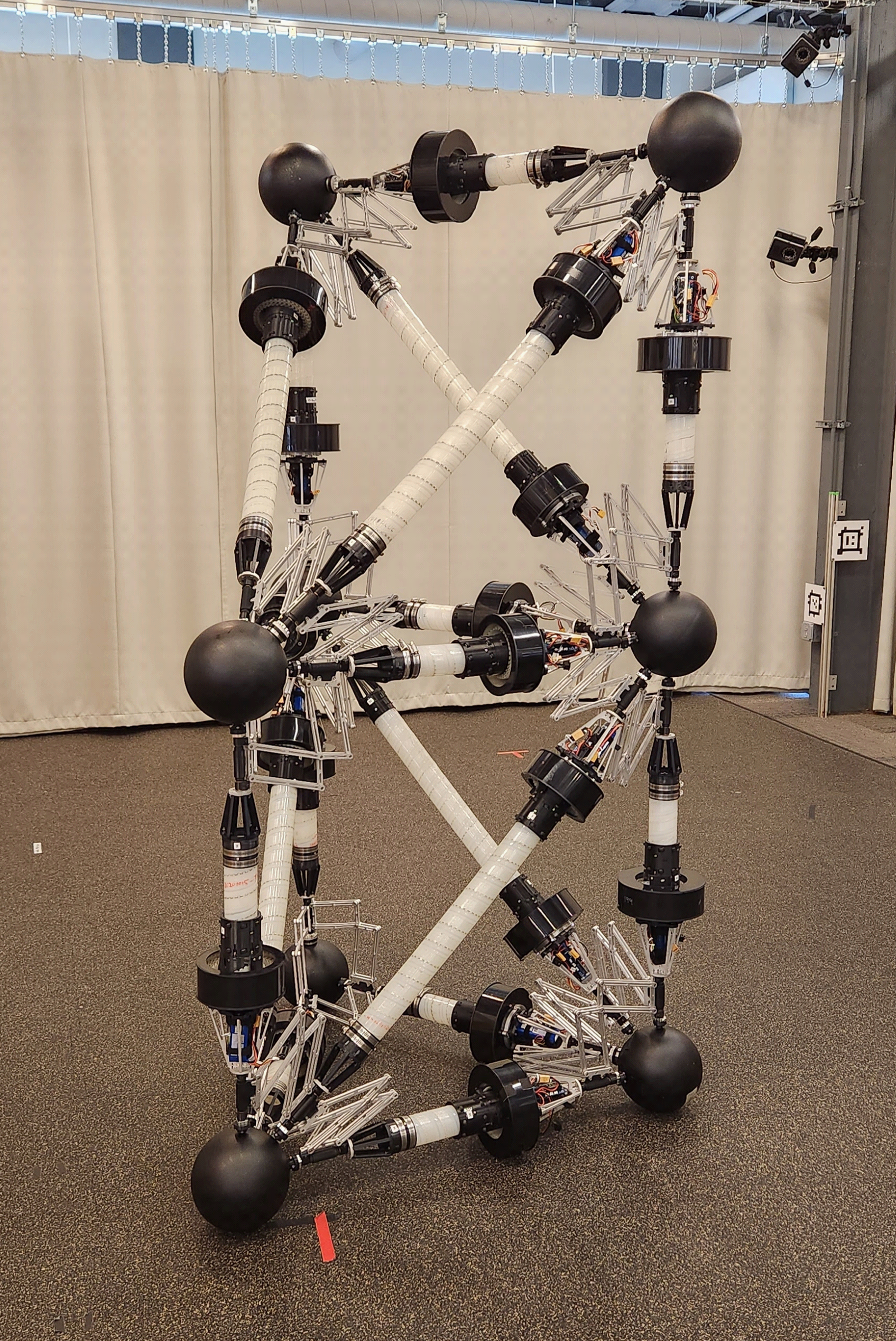}
    \caption{}
    \label{Fig_1_c}
  \end{subfigure}
\end{minipage}
\hfill
\begin{minipage}[t]{0.41\textwidth}\vspace{0pt}\centering
  \begin{subfigure}[t]{\linewidth}
    \includegraphics[width=\linewidth]{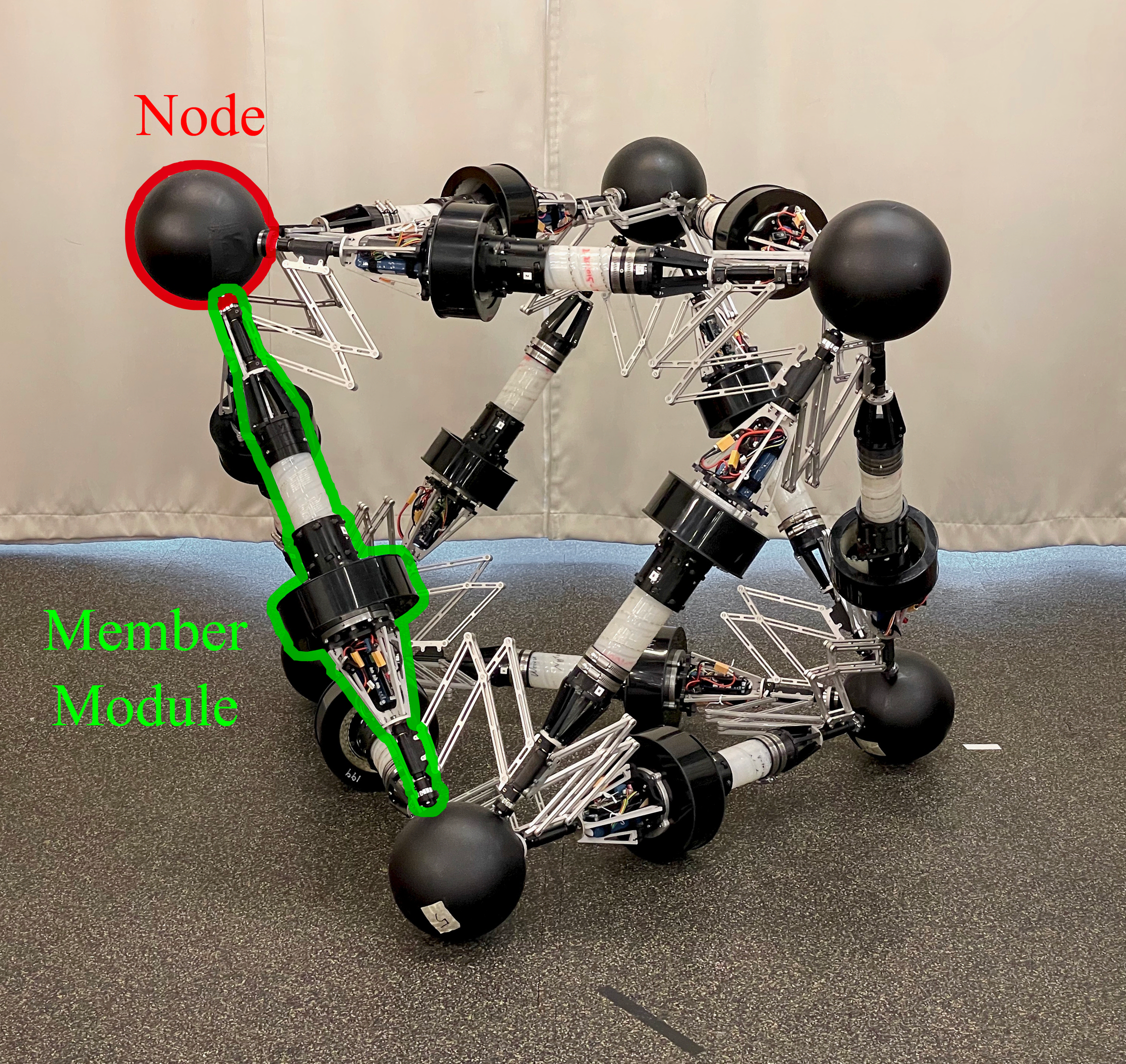}
    \caption{}
    \label{Fig_1_d}
  \end{subfigure}
\end{minipage}

\caption{The conceptual rendering and hardware prototype of the VTT system; (a) 3D-rendered scene of search-and-rescue operation of the VTT system; (b) Tent configuration of the VTT system; (c) Tower configuration of the VTT system; (d) Octahedron configuration of the VTT system.}
\label{Fig_VTTIntro}
\end{figure}

The Variable Topology Truss (VTT) system is a truss robotic system that includes the capability to self-reconfigure its topology \cite{spinos2021topological}.
The original goal was to deploy in unstructured environments, such as damaged buildings and provide structural support for search and rescue operations in case of collapse, Figure~\ref{Fig_1_a}.
By using spiral zippers as actuators for the truss member modules, the VTT system can achieve significant size changes that were not possible in previous truss robot systems \cite{collins2016design, lee2021slip}.
To enable effective search-and-rescue operations,  locomotion planning is part of the  task sequence \cite{park2019optimization, park2020polygon, bae2021polygon, bae2023locomotion}.
Other important tasks that can take advantage of the VTT system's ability to reconfigure its topology \cite{spinos2017towards}.
Object manipulation tasks in the search and rescue context include delivering medical supplies or removing debris.
One challenge for VTT for object manipulation comes from the spiral zipper's high friction and the VTT system's complex kinematics and statics which leads to difficulty developing a controller that can track VTT contact point position and force \cite{lee2021slip}.

This paper presents an object manipulation strategy for the VTT system based on a hierarchical hybrid position/force control.
First, we develop force controllers at both the actuator level and the task-space level.
Because the spiral zipper actuator has high friction, each actuator uses sensor-based force feedback to generate the desired axial force.
Based on the static model of the VTT system, we formulate a method to compute the member forces required to produce a desired task-space force at the target nodes.
The force tracking performance of a single actuator and the VTT system are tested with multiple experiments.
Two representative configurations are examined: a tetrahedral topology, representing the simplest fully-constrained structure, and a pyramid topology, which is the simplest over-constrained system. 
Force control experimental results show that the force tracking performance is reliable for both fully and over-constrained truss structures.

Object manipulation requires simultaneous regulation of position and force.
Therefore, we propose a hybrid control strategy that achieves concurrent position and force control without explicit decoupling.
This approach enables the VTT system to simultaneously track position and force along the same direction, which is not possible with decoupled control.
We proposed and experimentally validated a manipulation method that using two VTT nodes to grasp and transport an object.
The manipulation performance of the VTT system is validated with two different configurations, double tetrahedron and octahedron with internal nodes.
Experimental results show that the hybrid controller achieves consistent and reliable manipulation for both cases.
Given the VTT's intended role in real-world disaster response, the proposed hierarchical hybrid control framework forms a foundation for autonomous and adaptive manipulation in unstructured environments.

\section{Materials and Methods}

\subsection{Variable Topology Truss System}

The VTT system is a truss robot system that can drastically change its size and reconfigure its topology.
As shown in Figure~\ref{Fig_VTTIntro}, the VTT system can achieve various configurations by assembling the truss modules in different ways.
A {\em member module} is a truss element of the VTT system.
A {\em node} is a passive spherical joint that connects member modules.
In Figure~\ref{Fig_1_d}, a member module is highlighted in green, while a node is highlighted in red.

\subsubsection{Member Module}

\begin{figure}[t]
    \centering
    \includegraphics[width=3.2in]{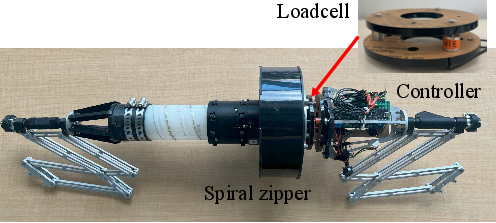}
    \caption{Picture of a member module in the VTT system}
    \label{Fig_MemberModule}
\end{figure}

Each member module is equipped with a linear actuator, sensors, a controller, and a battery.
Therefore, each module can operate independently and be easily assembled in different configuration without interfering with other modules. 
A loadcell assembly is attached for force sensing of each truss element of the VTT system.
Figure~\ref{Fig_MemberModule} presents a photograph of an individual member module.

\begin{figure}[t]
    \centering
    \includegraphics[width=3.2in]{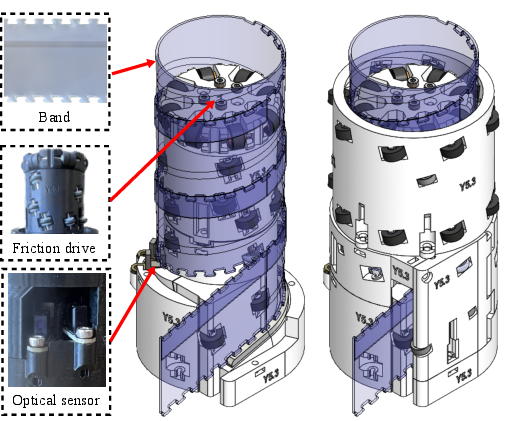}
    \caption{The CAD modeling of the spiral zipper actuator}
    \label{Fig_SpiralZipper}
\end{figure}

Each member module includes a spiral zipper actuator for linear actuation.
The spiral zipper actuator is a linear actuator capable of achieving an exceptionally large extension ratio \cite{collins2016design}.
The spiral zipper operates by winding up a long flat band (0.05'' thick nylon), which helps minimize the actuator's retracted length.
For the VTT system, the spiral zipper design has been modified to ensure compatibility with the VTT system, as shown in Figure~\ref{Fig_SpiralZipper}.
A nylon band column is rotated by a motor (118 RPM HD Premium Planetary Gear Motor, ServoCity) in the center of the column, while the length of the actuator is measured with two photointerrupters (OPB628).
The two photointerrupters are positioned with a $90^\circ$ phase offset to detect the direction of band motion.

\subsubsection{Control System Summary}

\begin{figure}[t]
    \centering
    \includegraphics[width=3.2in]{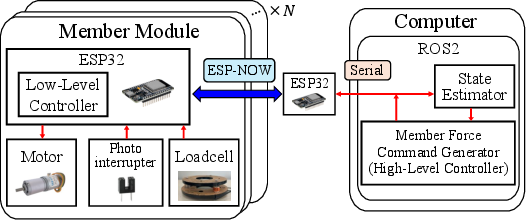}
    \caption{The control architecture of the VTT system}
    \label{Fig_ControlSummary}
\end{figure}

The control architecture of the VTT system is summarized in Figure~\ref{Fig_ControlSummary}. 
Each member module uses an ESP32 microcontroller (ESP32-PICO-KIT v4) that executes the low-level control loop. 
Using loadcell feedback, the low-level controller regulates the actuator to produce the commanded axial force. 
The controller also estimates the member length using two photointerrupters by counting the teeth on the spiral zipper band passing the sensor.
The ESP32s on the member modules communicate wirelessly using the ESP-NOW protocol with a master ESP32 connected to a host computer. The ESP-NOW protocol provides lightweight, low-latency communication compared to standard Wi-Fi.

The high-level controller is implemented in ROS2 and runs on the host computer. 
It receives state information from each module via the master ESP32 and computes the desired force commands for the member modules based on the measured module states and the required position and force at the target node. 
These force commands are then transmitted back to the member modules through the master ESP32. 
The details of the high-level and low-level controllers are presented in the following sections.

\subsubsection{VTT Kinematic Analysis}

\begin{figure}[t]
    \centering
        \includegraphics[width=2.3in]{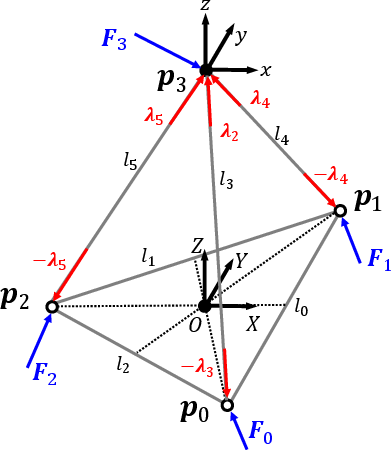}
    \caption{Mathematical schematics of the VTT system in a tetrahedral configuration. Black circles represent the nodes of the VTT, and gray lines represent the Spiral-Zipper beams.}
    \label{Fig_kinematics}
\end{figure}

The length between two connected nodes $(l_i)$ can be expressed as follows from Figure~\ref{Fig_kinematics}: 
\begin{equation}\label{length_definition}
l_i = \left| (\bm{p}_{i} - \bm{p}_{j}) \right|,
\end{equation}
where $\bm{p}_{i}=[p_{ix}, p_{iy}, p_{iz}]^\top$ denotes the position vector of $i$-th node. 

We define the stacked node position vector $\bm{P}=[{\bm{p}_{1}}^\top,\cdots,{\bm{p}_{N}}^\top]^\top$ and the member length vector $\bm{L}=[l_1,\cdots,l_M]^\top$, where $N$ and $M$ denote the numbers of nodes and members.
By combining the relation between node positions and member lengths in VTT can be expressed as follows.:
\begin{equation}\label{length_definition_total}
\bm{L} = f(\bm{P}),
\end{equation}
where $f(x)$ is a function  derived from Eq.~\ref{length_definition}. 
By differentiating Eq.~\ref{length_definition_total} with respect to node position, the inverse kinematics equation can be found from  Figure~\ref{Fig_kinematics}:

\begin{equation}\label{general_kinematics}
\dot{\bm{L}} = \mathbf{J}\dot{\bm{P}},
\end{equation} 
where matrix $\mathbf{J}$ is the inverse Jacobian of the VTT, mapping node velocities to member length velocities.

\subsubsection{VTT Static Analysis}

Based on  previous studies, the motion of the VTT system can be considered  quasi-static.
The relation between the internal forces and the nodal forces can be derived using the transpose of the inverse Jacobian matrix.

\begin{equation}\label{static_analysis}
  \bm{F} = \mathbf{J}^\top \bm{\lambda} + \bm{W},
\end{equation}
where $\bm{F} = [{\bm{F}_1}^\top, \cdots, {\bm{F}_N}^\top]^\top$ is the stacked vector of nodal forces,  while $\bm{\lambda} = [\lambda_1, \cdots, \lambda_M]^\top$ denotes the vector of internal forces.
$\mathbf{J}$ is the inverse Jacobian matrix derived in Eq.~\ref{general_kinematics}.
$\bm{W} =[{\bm{w}_1}^\top, \cdots, {\bm{w}_N}^\top]^\top$ is the vector of the external forces applied to each node includes gravitational forces.

Let $E_k$ denote the set of member indices connected to node $k$.
While the index $i \in E_k$ denotes the indices of the members that are connected to the node $k$, the required member force vector $\bm{\lambda}$ to generate the desired force $\bm{F}_{des}$ at node $k$ can be calculated as follows:
\begin{equation}\label{force_calculation}
  \bm{\lambda}_k = \left({\mathbf{J}_k}^\top\right)^\dagger \left(\bm{F}_{des} - \bm{W}_k\right)
\end{equation}
where $\bm{\lambda}_k$ is the vector of member forces $\lambda_i$, which is the forces of the members connected to the node $k$.
$\mathbf{J}_k$ is the submatrix of the inverse Jacobian $\mathbf{J}$ formed by selecting the columns corresponding to $i \in E_k$.
$\bm{W}_k$ denotes the external force vector applied to node $k$.
The symbol $\dagger$ denotes the Moore-Penrose pseudoinverse.

\subsection{Hierarchical Hybrid Control of the VTT System}

\begin{figure}[t]
  \centering
  \includegraphics[width=0.48\textwidth]{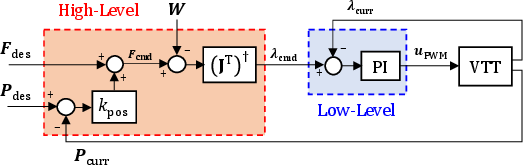}
  \caption{Block diagram of the hybrid position/force control for object manipulation using the VTT system.}
  \label{Fig_block_diagram}
\end{figure}

To perform grasping and manipulation with the VTT system, the nodal position and contact force must be controlled simultaneously.
We propose a hierarchical hybrid controller that achieves concurrent position and force control without explicit decoupling, enabling tracking of position and force along the same direction.
Figure~\ref{Fig_block_diagram} presents the overall control structure.
The high-level controller computes the desired nodal force command to track both position and force and maps this command to the corresponding member force commands using the system statics. 
The low-level controller then drives each actuator to generate the required member forces received from the high-level controller.

\subsubsection{High-Level Controller}
The high-level controller generates a force command for the target node to simultaneously track the desired force and position.
The commanded nodal force is computed by summing the desired force and a proportional feedback term based on the position error, as follows:
\begin{equation}
    \bm{F}_{\mathrm{cmd}} = \bm{F}_{\mathrm{des}} + k_{\mathrm{pos}} \left( \bm{P}_{\mathrm{des}} - \bm{P}_{\mathrm{curr}} \right),
\label{eq_control_command}
\end{equation}
where $\bm{F}_{\mathrm{des}}$ is the stacked vector of desired nodal forces applied to the environment.
$\bm{P}_{\mathrm{des}}$ denotes the stacked desired node positions, and $\bm{P}_{\mathrm{curr}}$ is the stacked current node positions.
$k_{\mathrm{pos}}$ is the proportional gain for position control part which can be interpreted as the stiffness gain that converts end-effector node position deviations into equivalent stiffness forces.
The member forces command $\bm{\lambda}_\mathrm{cmd}$ required to generate the nodal force $\bm{F}_{\mathrm{cmd}}$ can be derived from Eq.~\ref{force_calculation} by substituting the command force vector $\bm{F}_{\mathrm{cmd}}$ into $\bm{F}_{\mathrm{des}}$, and the weight of nodes and members compensation vector $\bm{W}$ into $\bm{W}_k$.
$\bm{\lambda}_\mathrm{cmd}$ is the stacked vector of a member force command $\lambda_{\mathrm{cmd}, i}$ for each member.
This way, the control law generates an equivalent force to tracking position and force simultaneously, enabling high level unified force-based command synthesis.

\subsubsection{Low-Level Controller}

The low-level control directly interfaces with VTT's actuation system to execute the high-level force commands. 
A Proportional-Integral (PI) controller is implemented to regulate the actuator output. 
The controller can be expressed as follows:
\begin{equation}\label{modeling_eq}
    u_{\mathrm{PWM}} = k_P e_{\lambda} + k_I \int e_{\lambda} dt,
\end{equation}
where $u_{\mathrm{PWM}}$ is the pulse width modulation (PWM) input for controlling the motor.
$e_{\lambda} = \lambda_{\mathrm{cmd}} - \lambda_{\mathrm{curr}}$ is the error between the desired member force $\lambda_{\mathrm{cmd}}$, derived from the high-level controller, and the current member force $\lambda_{\mathrm{curr}}$, measured from the loadcell assembly.

\subsubsection{Remark on Decoupling and Stability}

Most position-force hybrid controllers decouple the force and position directions into orthogonal subspaces using selection operators \cite{an1989role, fisher1992hybrid}.
In contrast, by avoiding this decoupling, the system can control force and position along the same direction simultaneously, but with the cost of increased tracking error.
Assuming quasi-static motion, each channel reduces to a proportional or impedance-type feedback loop.
Therefore, all control components have exponentially stable error dynamics in its respective subspace for positive gains. 
By adjusting the stiffness gain $k_f$, we can adjust control strength between position and force to keep the closed-loop tracking errors  bounded. 
Since this study focuses on demonstrating experimental feasibility, we select $k_f$ to prioritize position tracking accuracy over force tracking performance.

\begin{figure}[t]
\centering

\begin{minipage}[t]{0.55\textwidth}
  \centering
  \begin{subfigure}[t]{0.9\linewidth}
    \includegraphics[width=0.9\linewidth]{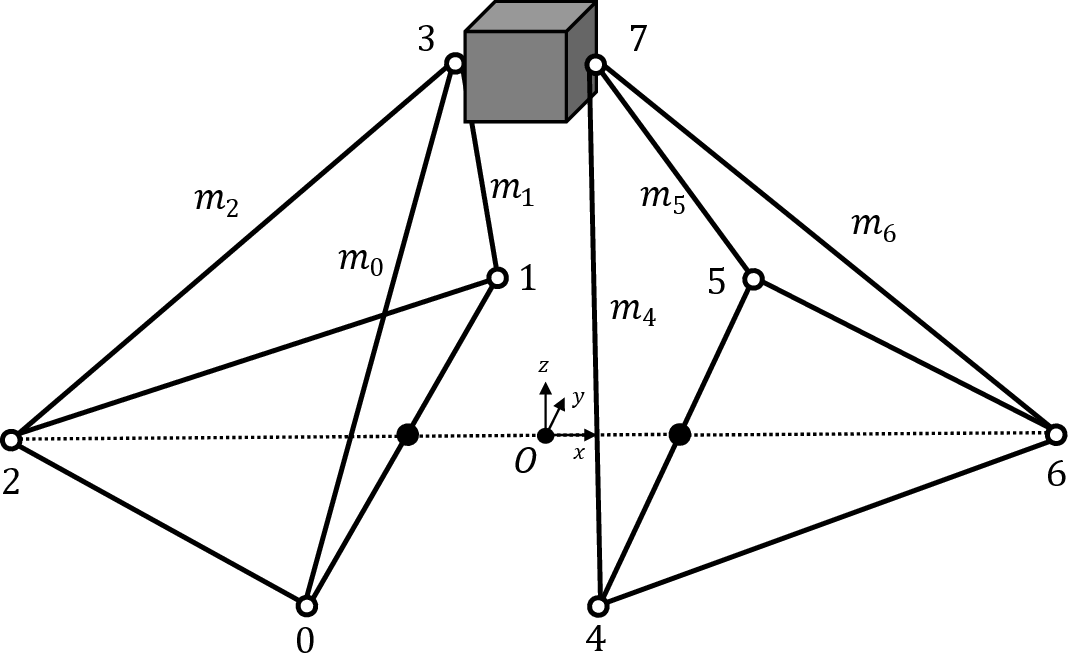}
    \caption{}
    \label{Fig_Manipulation_a}
  \end{subfigure}
\end{minipage}
\hspace{0.04\textwidth}
\begin{minipage}[t]{0.4\textwidth}
  \centering
  \begin{subfigure}[t]{\linewidth}
    \includegraphics[width=\linewidth]{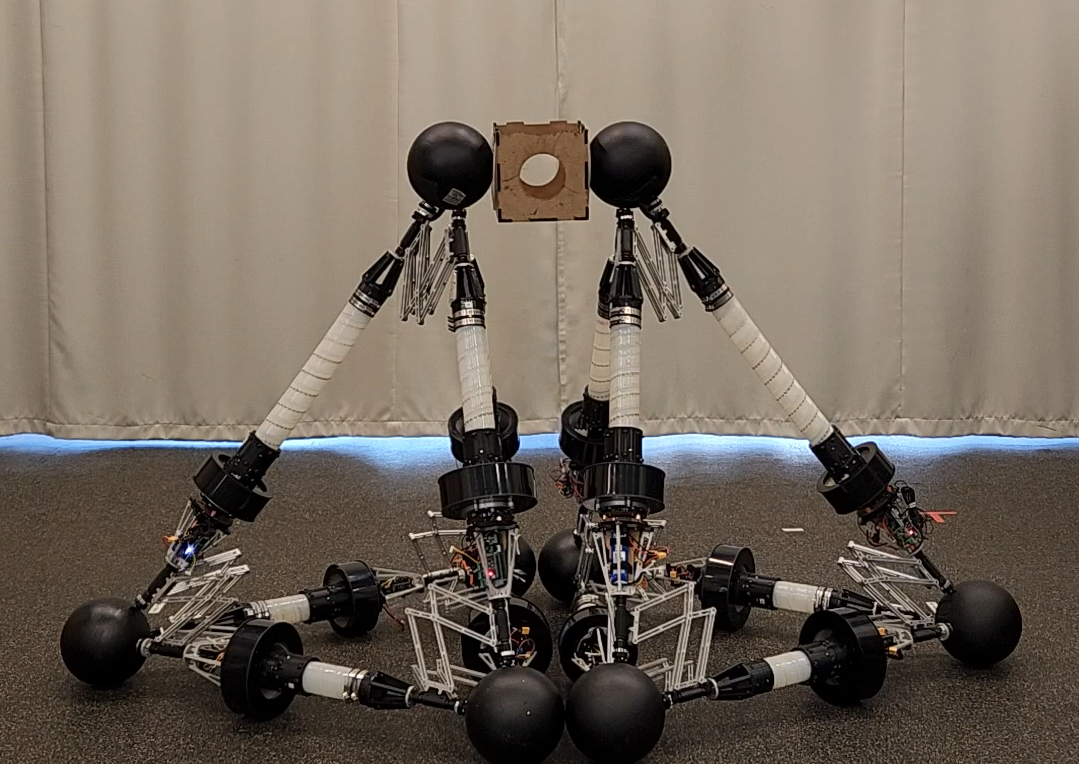}
    \caption{}
    \label{Fig_Manipulation_b}
  \end{subfigure}
\end{minipage}

\vspace{6pt}

\begin{minipage}[t]{0.6\textwidth}
  \centering
  \begin{subfigure}[t]{\linewidth}
    \includegraphics[width=\linewidth]{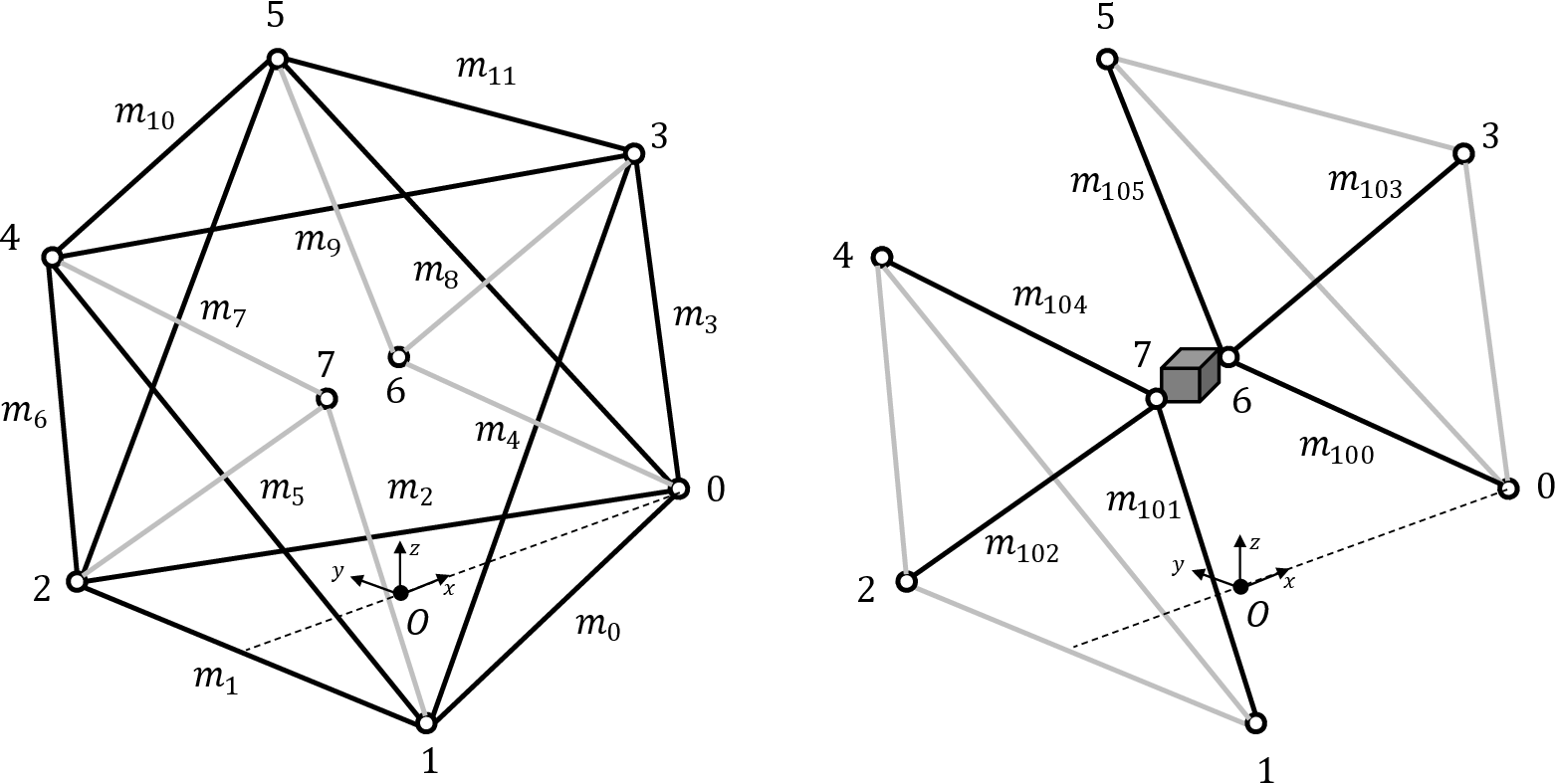}
    \caption{}
    \label{Fig_Manipulation_c}
  \end{subfigure}
\end{minipage}
\hspace{0.04\textwidth}
\begin{minipage}[t]{0.35\textwidth}
  \centering
  \begin{subfigure}[t]{\linewidth}
    \includegraphics[width=\linewidth]{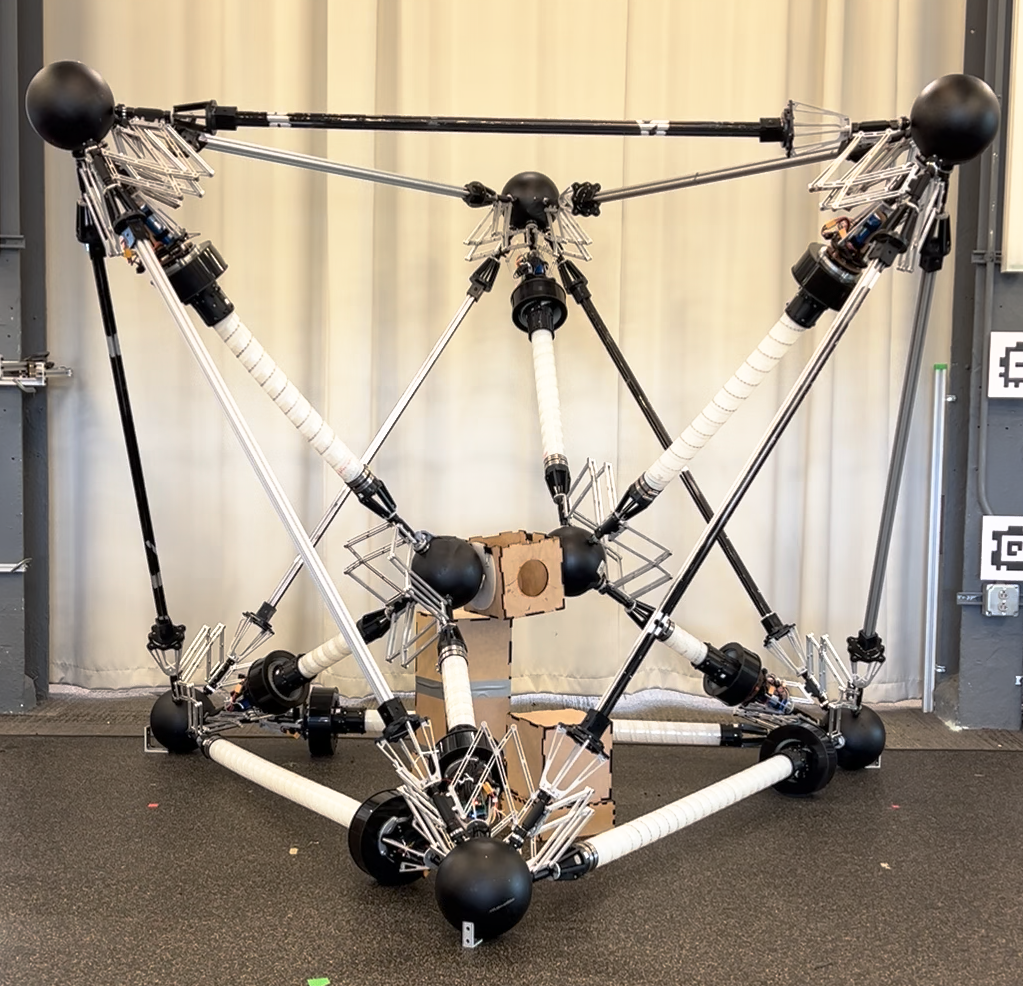}
    \caption{}
    \label{Fig_Manipulation_d}
  \end{subfigure}
\end{minipage}

\caption{Two manipulation configuration of VTT. (a) Schematic of the double tetrahedral configuration; (b) Picture of the double tetrahedral configuration; (c) Schematic of the octahedral configuration; (d) Picture of the octahedral configuration.}
\label{Fig_Manipulation}
\end{figure}

\subsection{Object Manipulation of the VTT System}

In addition to controlling the position of end effectors in contact with an object we would like to control internal forces on the object to ensure a positive grasp. We propose a hierarchical hybrid control  for  object manipulation  with the VTT system. 
This way the VTT system can control the object that results in both position as well as static controlled forces.
To verify the objective manipulation performance, we propose two configurations of VTT.

The double-tetrahedron configuration is the most intuitive and simplest setup for object manipulation.
It consists of two identical tetrahedral VTTs. 
The system manipulates an object by pushing it with the two top nodes, as shown in Figures~\ref{Fig_Manipulation_a} and~\ref{Fig_Manipulation_b}.
Although the double-tetrahedron configuration is easy to control, it has several limitations, such as a restricted workspace and difficulty representing realistic manipulation scenarios.
Therefore, the double-tetrahedron configuration is used for detailed data generation. 
Multiple trajectories were executed with this configuration, and the resulting position and force data were analyzed.

The octahedron configuration uses two internal tetrahedrons to manipulate a target object, as shown in Figures~\ref{Fig_Manipulation_c} and~\ref{Fig_Manipulation_d}.
The two internal nodes press the object from opposite sides to hold it in place.
Because this configuration better represents realistic manipulation scenarios, we used it to test practical object manipulation tasks.

\section{Results}

The control performance of the hybrid controller is evaluated experimentally before testing manipulation.
First, a force-tracking experiment on a single spiral zipper actuator is conducted to verify actuator-level force control performance.
Next, we perform force-tracking experiments with two different VTT configurations to assess whether the node-level force control performance is sufficient for manipulation.

After validating the force-tracking performance, we evaluate manipulation performance using both the double-tetrahedron and octahedron configurations.
In the double-tetrahedron configuration, we primarily test combined position and force tracking over multiple trajectories.
In the octahedron configuration, the manipulation task is designed as a more practical, application-oriented scenario.

\begin{figure}[t]
    \centering
    \begin{subfigure}{0.32\textwidth}
        \centering
        \includegraphics[height=5cm, valign=c]{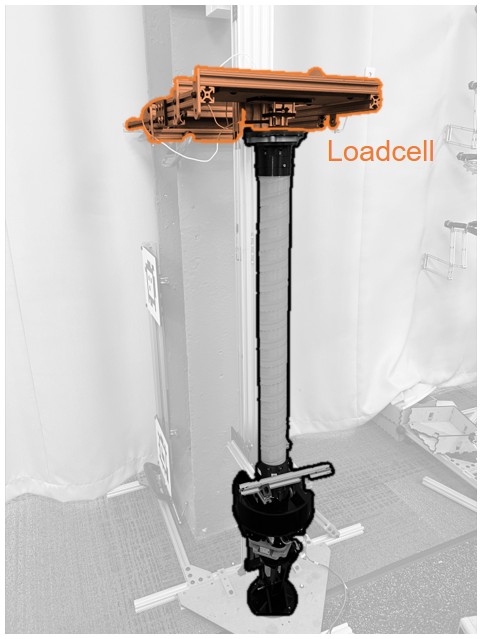}
        \caption{}
        \label{Fig_spiral_zipper_testbench}
    \end{subfigure}
    \begin{subfigure}{0.32\textwidth}
        \centering
        \includegraphics[height=5cm, valign=c]{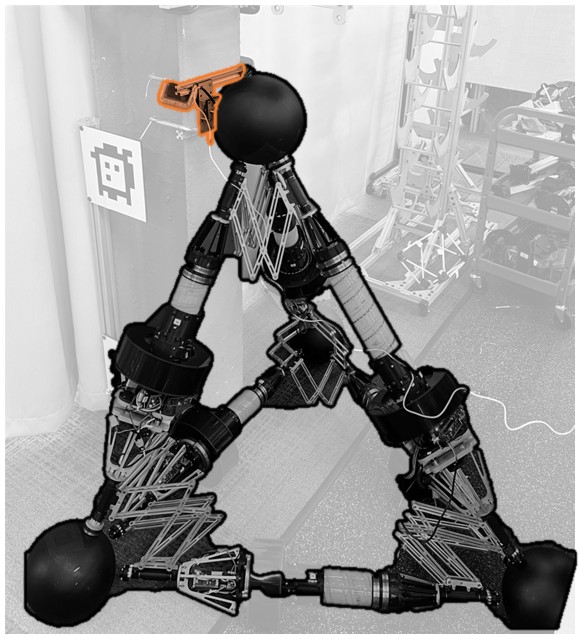}
        \caption{}
        \label{Fig_tetra_testbench}
    \end{subfigure}
    \begin{subfigure}{0.32\textwidth}
        \centering
        \includegraphics[height=5cm, valign=c]{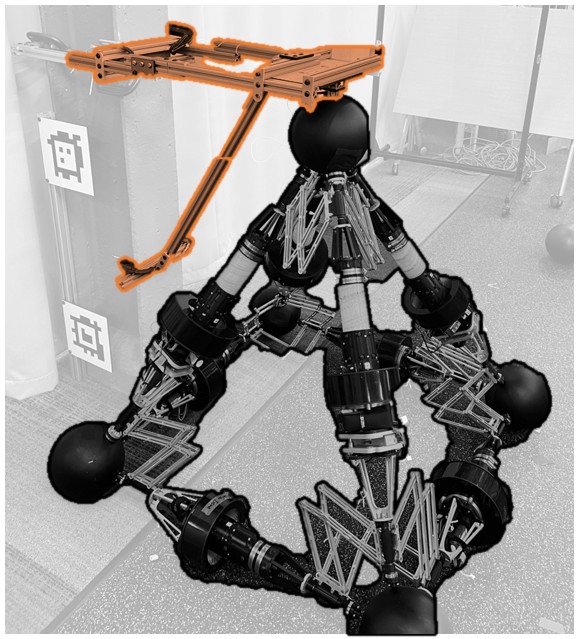}
        \caption{}
        \label{Fig_pyramid_testbench}
    \end{subfigure}
    \caption{Experimental setups for force control evaluation. Orange components indicate the force measurement devices equipped with load cell. (a) Testbench for Spiral Zipper; (b) Testbench for tetrahedron topology VTT; (c) Testbench for pyramid topology VTT.}
    \label{Fig_exp_setup}
\end{figure}

\subsection{Force Control of a Single Spiral Zipper}

To measure the force applied by a spiral zipper actuator, we constructed a dedicated test fixture, as shown in Figure~\ref{Fig_spiral_zipper_testbench}.
The test fixture consists of a fixed vertical mounting structure and a load cell at the free end to measure compressive forces.
We used a force signal that increases at \SI{10}{N/s} until it reaches the desired maximum to avoid unnecessary oscillations.
The maximum values of the ramp inputs were set between \SI{10}{N} and \SI{200}{N}.

\begin{figure}[t]
    \centering
    \begin{subfigure}{0.35\textwidth}
        \centering
        \includegraphics[width=\linewidth]{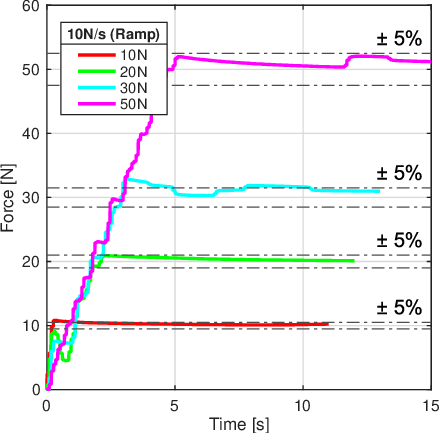}
        \caption{}
     \label{10_sz10_50}
    \end{subfigure}
    \hspace{3mm}
    \begin{subfigure}{0.35\textwidth}
        \centering
        \includegraphics[width=\linewidth]{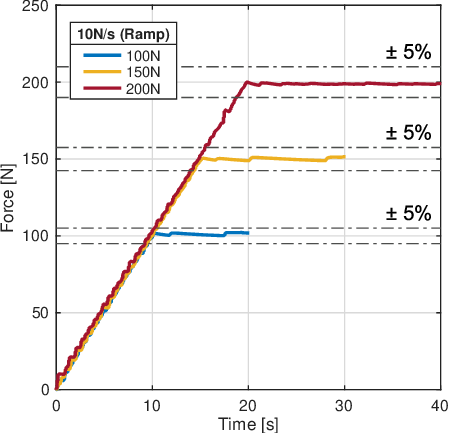}
        \caption{}
     \label{10_sz100_200}
    \end{subfigure}

\caption{Experimental results of force control of the Spiral-Zipper under ramp
input signals of \(10\,\mathrm{N/s}\).
(a) Force responses in the range of \(10\text{–}50\,\mathrm{N}\).(b) Force responses in the range of \(100\text{–}200\,\mathrm{N}\).}
\label{10_SZ_exp}
\end{figure}

The force response results are summarized in Figure~\ref{10_SZ_exp}.
During the force-ramping phase, the output force exhibits a steeper slope at lower target forces, despite the constant input rate (Figure~\ref{10_sz10_50}).
Minor errors appear only in specific low-force regions, caused by the dead zone of the spiral zipper actuator. 
The dead zone is caused by design characteristics of the spiral zipper's internal friction-drive mechanism, which requires a relatively high actuation torque to initiate motion.
This dead zone prevents motion under low input signals, resulting in step-like behavior in the low-force range.
During the force-holding phase, the module consistently sustains the desired force with less than 5\% error across all tested ranges. 
These results confirm that the spiral zipper actuator can reliably generate and maintain axial forces, demonstrating its suitability for integration into the VTT system.

\subsection{Force Control of the VTT System}

We could not find any previous studies that conducted an experimental validation of force-control performance for VGT or similar truss robot systems.
Most prior studies on manipulation have relied solely on simulation to evaluate motion and force performance.
Several studies have experimentally demonstrated force control only at the single-actuator level \cite{rost2012design}.
For evaluating the force-tracking performance of the full VTT system, external load cells were mounted near the target node to measure the applied forces, as shown in Figure~\ref{Fig_tetra_testbench} and~\ref{Fig_pyramid_testbench}.
We selected tetrahedron and pyramid configurations as reference cases.
The tetrahedron configuration is the simplest rigid, fully constrained truss structure, while the pyramid configuration allows us to assess force control performance in an over-constrained truss system.
We used  similar force signals used for a single spiral zipper testing, the force signal  increases at \SI{10}{N/s} until it reaches the desired maximum value.
The maximum values of the force inputs were set between \SI{10}{N} and \SI{50}{N}.
We tested force generation in both the horizontal ($x$) and vertical ($z$) directions.

\subsubsection{Tetrahedral Configuration}

The results of the force-tracking experiments in the tetrahedral configuration are shown in Figure~\ref{Fig_topology_Force}. 
In both the horizontal and vertical directions, the applied force remains within 5\% of the target value during the force-holding phase.

For the $x$-direction, a significant overshoot is observed at lower target forces as shown in Figure~\ref{Fig_Tet_Force_x}. 
We attribute this behavior to the combined effects of the actuator dead zone and friction in the node joints. 
Here, the spiral zipper actuators begin to move abruptly only after the input current exceeds a certain threshold, which leads to  overshoot.

In the $z$-direction, the overall performance was more stable, with smaller errors and less overshoot, as shown in Figure~\ref{Fig_Tet_Force_z}. 
The observed step-like behavior is also caused by the dead zone of the spiral zipper actuator, particularly under increasing compressive loads. 
Despite these behaviors, the VTT system in the tetrahedral configuration demonstrated reliable force-tracking performance in both directions, showing the effectiveness of the proposed control method in a fully constrained system.

\subsubsection{Pyramid Configuration}

\begin{figure}[t]
    \centering
    \begin{subfigure}{0.35\textwidth}
        \centering
        \includegraphics[width=\linewidth]{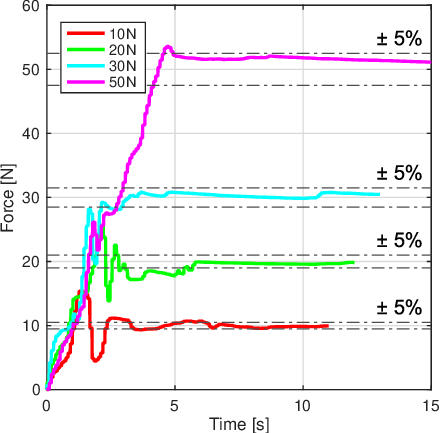}
        \caption{}
        \label{Fig_Tet_Force_x}
    \end{subfigure}
    \hspace{3mm}
    \begin{subfigure}{0.35\textwidth}
        \centering
        \includegraphics[width=\linewidth]{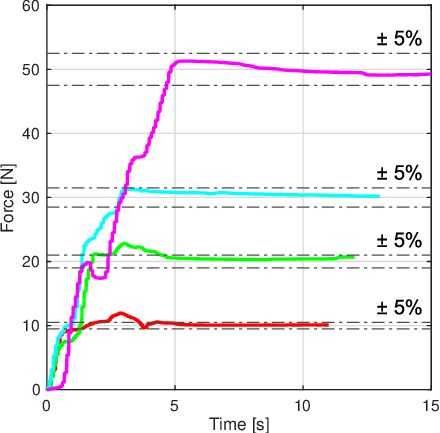}
        \caption{}
        \label{Fig_Tet_Force_z}
    \end{subfigure}

    \vspace{2mm}

    \begin{subfigure}{0.35\textwidth}
        \centering
         \includegraphics[width=\linewidth]{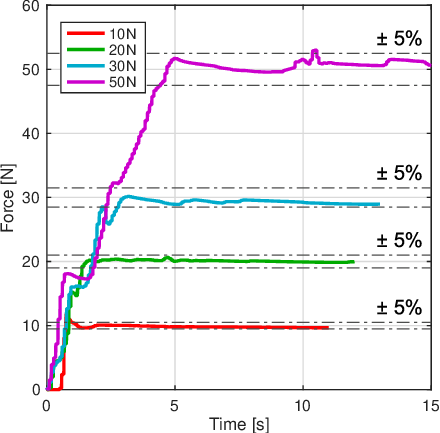}
        \caption{}
        \label{Fig_Py_Force_x}
    \end{subfigure}
    \hspace{3mm}
    \begin{subfigure}{0.35\textwidth}
        \centering
        \includegraphics[width=\linewidth]{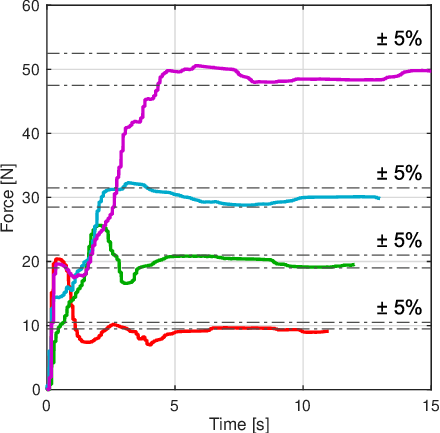}
        \caption{}
        \label{Fig_Py_Force_z}
    \end{subfigure}

    \caption{Experimental results of force control in the tetrahedral topology under target forces ranging from \(10\text{–}50\,\mathrm{N}\). 
    (a)–(b) Force responses for tetrahedron topology along the \(x\)- and \(z\)-axes. 
    (c)–(d) Force responses for pyramid topology along the \(x\)- and \(z\)-axes.}
    \label{Fig_topology_Force}
\end{figure}

The experimental results of force tracking in the pyramid topology are shown in Figure~\ref{Fig_topology_Force}. 
This configuration is selected to evaluate force control performance in an over-constrained structure. 
In both directions, the error during the force-holding phase remains within approximately 5\% across all target force levels, indicating overall stable performance.

Similar to the tetrahedron configuration results, a significant overshoot is observed, particularly in the $z$ direction. 
The overshoot-related errors are larger in the pyramid configuration, which we attribute to the combined effects of the spiral zipper actuator dead zone and internal forces caused by the over-constrained structure. 
We considered the overshoot observed in the pyramid configuration to be acceptable given the over-constrained nature of the structure.
Therefore, we conclude that the proposed hybrid position/force control method is successful.

\subsection{Object Manipulation}

\subsubsection{Double Tetrahedral Configuration}

We could not find any prior studies have investigated two-node grasping with a VGT system.
Using the double-tetrahedron configuration, we evaluated the position and force tracking performance over the following trajectories:
\begin{itemize}
    \item Linear motion
    \begin{enumerate}
        \item x-axis motion between $+\SI{0.5}{m}$ and $-\SI{0.5}{m}$
        \item y-axis motion between $+\SI{0.35}{m}$ and $-\SI{0.35}{m}$
        \item y-axis motion from $0$ to $+\SI{0.5}{m}$ and back to the initial position
    \end{enumerate}
    \item Circular motion
    \begin{enumerate}
        \item Circular trajectory in  $xy$ plane (radius \SI{0.2}{m})
        \item Circular trajectory in  $yz$ plane (radius \SI{0.2}{m})
        \item Circular trajectory in  $xz$ plane (radius \SI{0.2}{m})
        \item Circular trajectory in  $xyz$ space (radius \SI{0.2}{m})
    \end{enumerate}
\end{itemize}

The target object is a wooden cube with dimensions $\SI{20}{cm} \times \SI{20}{cm} \times \SI{20}{cm}$ and a mass of \SI{931}{g}.
Silicone pads  attached to the sides of the object  increase the friction coefficient between the node spheres and the object.
The grasping force is set to \SI{60}{N} for all cases, which means that each grasping node applies \SI{30}{N} of force to the target object.
By using the hierarchical hybrid controller, the VTT system is controlled to follow the trajectory while maintaining the grasping force applied from  node 3 and node 7 in Figure~\ref{Fig_Manipulation_a}.

The position and force tracking performance data for all experiments are summarized in Table~\ref{Table_Linear} and Table~\ref{Table_Circular}.
Each experiment is repeated three times, and the root mean square error (RMSE) values are obtained by averaging the results of the three trials.
In all seven experiments, the position RMSE remains within  centimeters, which is reasonably accurate for manipulation.
The force RMSE is  larger compared to the commanded grasping force.
This is due to the gain settings of the hierarchical hybrid controller, because the controller does not decouple the position and force control inputs, there is a trade-off between position and force tracking performance with higher importance set to position accuracy.

\begin{figure}[t]
\centering
\begin{minipage}[b]{0.48\textwidth}
    \centering
    \includegraphics[width=\linewidth]{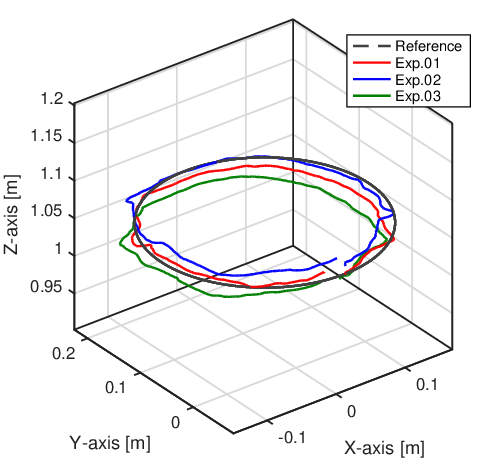}
    \caption*{(a)}

    \vspace{3mm}

    \includegraphics[width=\linewidth]{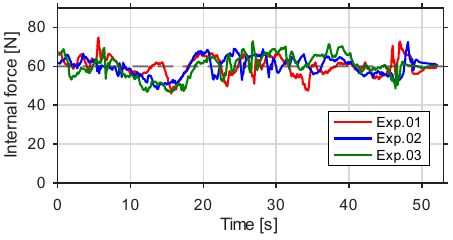}
    \caption*{(b)}
\end{minipage}
\hfill
\begin{minipage}[b]{0.48\textwidth}
    \centering
    \includegraphics[width=\linewidth]{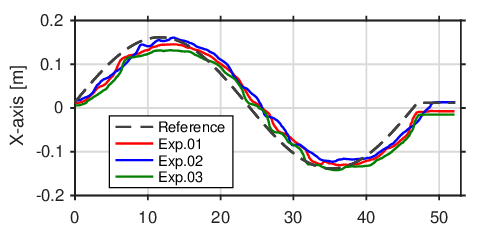}
    \caption*{(c)}

    \vspace{3mm}

    \includegraphics[width=\linewidth]{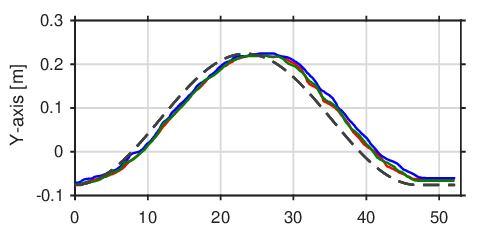}
    \caption*{(d)}

    \vspace{3mm}

    \includegraphics[width=\linewidth]{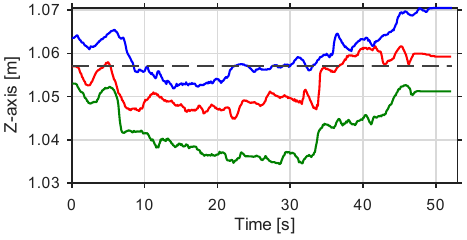}
    \caption*{(e)}
\end{minipage}
\caption{Experimental results of xy plane grab manipulation with two tetrahedron VTT. (a) 3D trajectory. (b) Internal force. (c) X-axis tracking result. (d) Y-axis tracking result. (e) Z-axis tracking result.}
\label{Fig_experimental_results_xy}
\end{figure}

\begin{figure}[t]
\centering
\begin{minipage}[b]{0.48\textwidth}
    \centering
    \includegraphics[width=\linewidth]{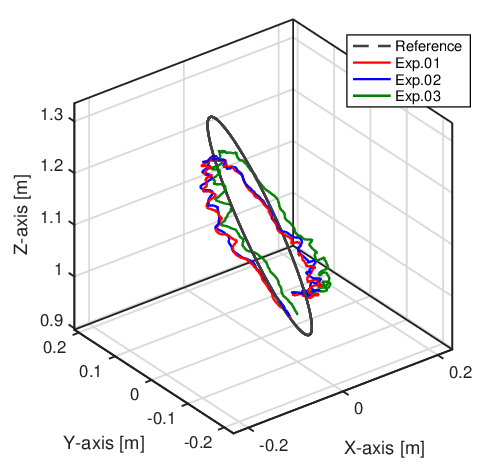}
    \caption*{(a)}

    \vspace{3mm}

    \includegraphics[width=\linewidth]{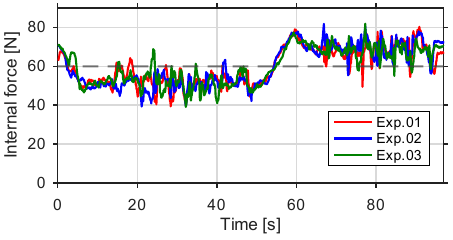}
    \caption*{(b)}
\end{minipage}
\hfill
\begin{minipage}[b]{0.48\textwidth}
    \centering
    \includegraphics[width=\linewidth]{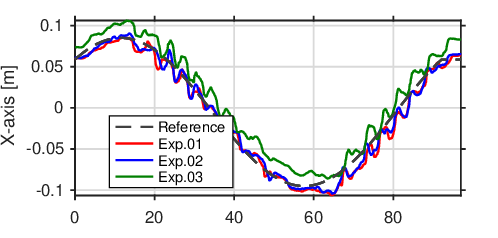}
    \caption*{(c)}

    \vspace{3mm}

    \includegraphics[width=\linewidth]{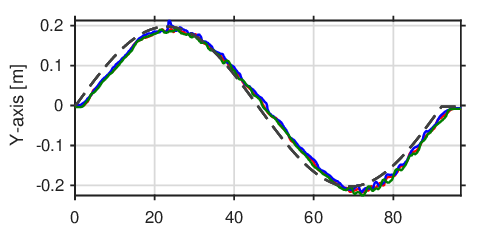}
    \caption*{(d)}

    \vspace{3mm}

    \includegraphics[width=\linewidth]{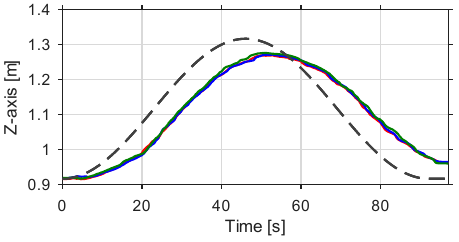}
    \caption*{(e)}
\end{minipage}
\caption{Experimental results of xyz plane grab manipulation with two tetrahedron VTT. (a) 3D trajectory. (b) Internal force. (c) X-axis tracking result. (d) Y-axis tracking result. (e) Z-axis tracking result.}
\label{Fig_experimental_results_xyz}
\end{figure}

Figure~\ref{Fig_experimental_results_xy} and $xyz$ plane in Figure~\ref{Fig_experimental_results_xyz} show detailed plots.~\ref{Fig_experimental_results_xy}, both the position and force data remain within a small error range for the $xy$ circular trajectory experiment.
The $xyz$ circular trajectory results indicate a phase delay in the $z$ position tracking.
In the case of $z$-position control, all three members must apply compressive or tensile forces toward each other.
This generates internal forces due to friction in the spherical joints at the nodes, reducing the effective force available for adjusting the $z$ position, which results in decreasing the maximum achievable $z$ displacement.
In addition, the spiral zipper actuator dead zone, adds positioning error.
The combination of these two issues results in a phase delay in the $z$-direction motion.

\begin{figure}[t]
\centering
\setlength{\tabcolsep}{0pt} 
\renewcommand{\arraystretch}{0} 

\newcommand{\Hgap}{0.02\textwidth}  
\newcommand{\Vgap}{4pt}             

\begin{tabular}{@{}c@{\hspace{\Hgap}}c@{}}

\begin{subfigure}[t]{0.35\textwidth}
  \includegraphics[width=\linewidth]{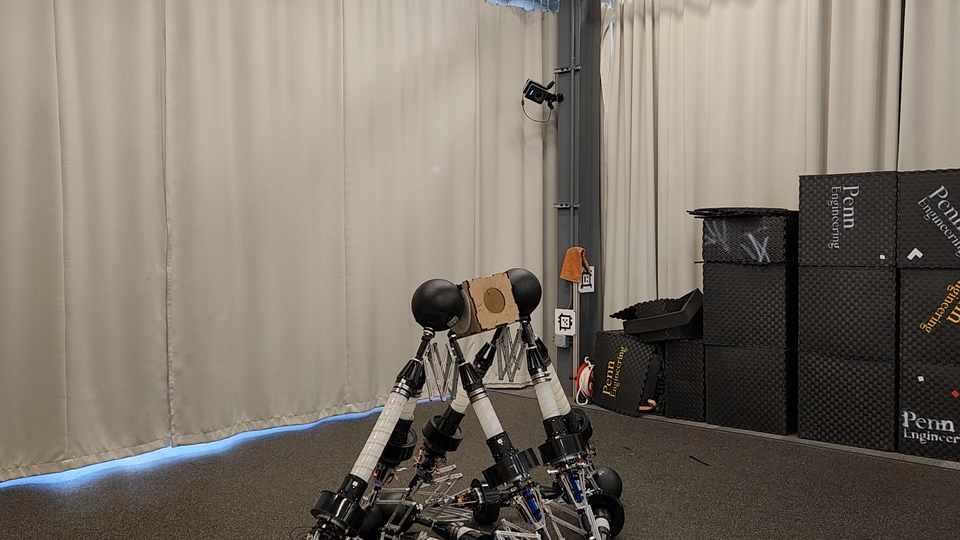}
  \caption{}
\end{subfigure}
&
\begin{subfigure}[t]{0.35\textwidth}
  \includegraphics[width=\linewidth]{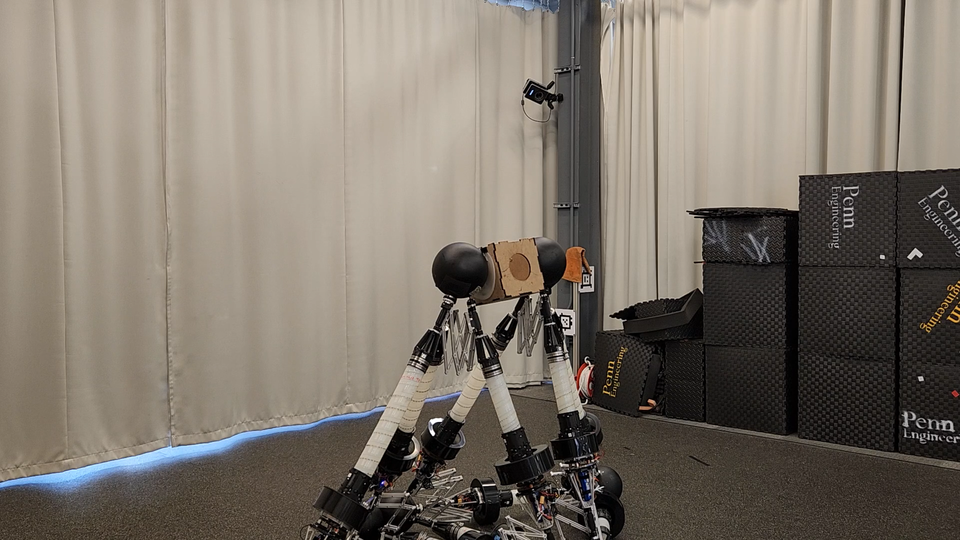}
  \caption{}
\end{subfigure}
\\[\Vgap]

\begin{subfigure}[t]{0.35\textwidth}
  \includegraphics[width=\linewidth]{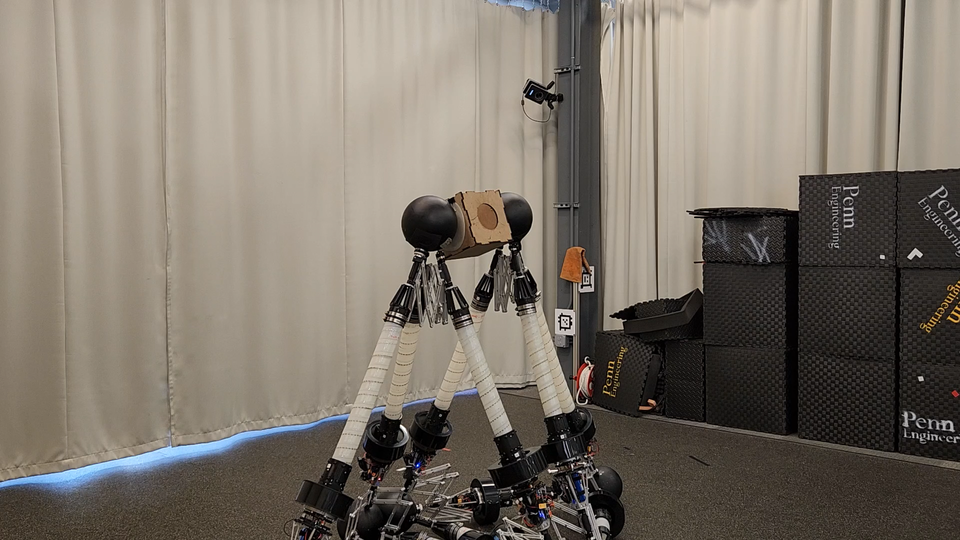}
  \caption{}
\end{subfigure}
&
\begin{subfigure}[t]{0.35\textwidth}
  \includegraphics[width=\linewidth]{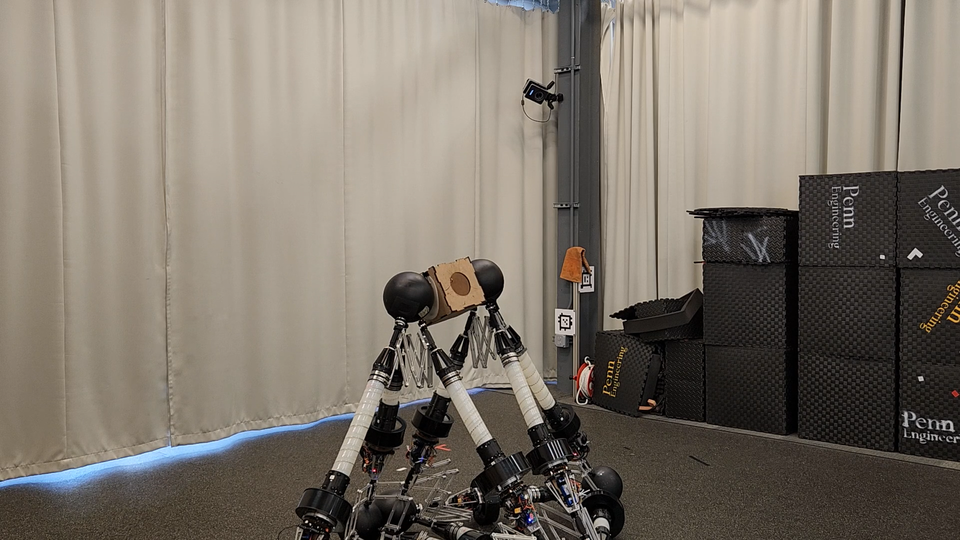}
  \caption{}
\end{subfigure}
\\[\Vgap]

\begin{subfigure}[t]{0.35\textwidth}
  \includegraphics[width=\linewidth]{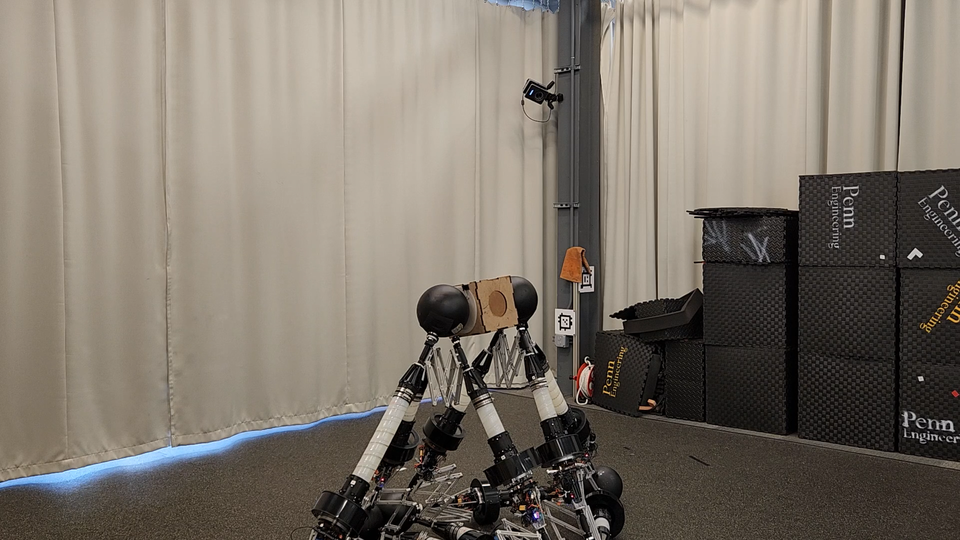}
  \caption{}
\end{subfigure}
&
\begin{subfigure}[t]{0.35\textwidth}
  \includegraphics[width=\linewidth]{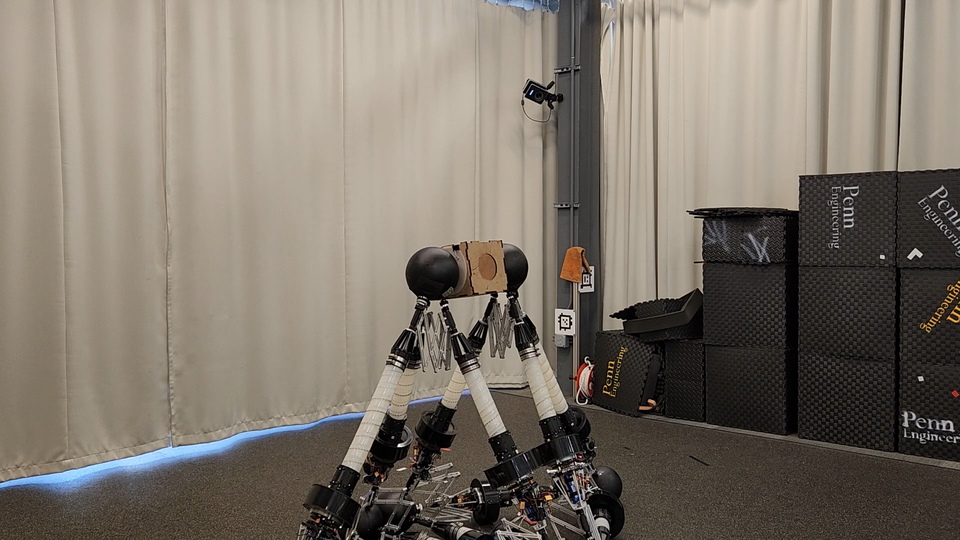}
  \caption{}
\end{subfigure}

\end{tabular}

\caption{Snapshots of the double tetrahedron VTT manipulation experiment. (a) 0:00~min; (b) 0:18~min; (c) 0:42~min; (d) 1:16~min; (e) 1:40~min; (f) 1:57~min.}
\label{Fig_DoubleTet_Exp}
\end{figure}

Figure~\ref{Fig_DoubleTet_Exp} shows the snapshots of the circular movement experiment on $xyz$ plane.

\begin{table}[htbp]
    \centering
    \small
    \setlength{\tabcolsep}{6pt}
    \renewcommand{\arraystretch}{1.1}
    \begin{threeparttable}
        \caption{Experimental results for line trajectories along different axes.}
        \label{Table_Linear}
        \begin{tabular*}{\linewidth}{@{\extracolsep{\fill}}l
                S[table-format=1.4]
                S[table-format=1.4]
                S[table-format=1.4]
                S[table-format=2.2]
            }
            \toprule
            {Axis} & {\(x\) (m)\tnote{a}} & {\(y\) (m)\tnote{b}} & {\(z\) (m)\tnote{c}} & {\(F\) (N)\tnote{d}} \\
            \midrule
            x-axis & 0.0321 & 0.0049 & 0.0197 & 7.30 \\
            y-axis & 0.0143 & 0.0421 & 0.0295 & 13.87 \\
            z-axis & 0.0118 & 0.0051 & 0.1372 & 5.67 \\
            \bottomrule
        \end{tabular*}
        \begin{tablenotes}
            \footnotesize
            \item[a] RMSE in \(x\)-direction.
            \item[b] RMSE in \(y\)-direction.
            \item[c] RMSE in \(z\)-direction.
            \item[d] RMSE of desired internal force.
        \end{tablenotes}
    \end{threeparttable}
\end{table}

\begin{table}[htbp]
    \centering
    \small
    \setlength{\tabcolsep}{6pt}
    \renewcommand{\arraystretch}{1.1}
    \begin{threeparttable}
        \caption{Experimental results for the circular trajectory executed on different planes.}
        \label{Table_Circular}
        \begin{tabular*}{\linewidth}{@{\extracolsep{\fill}}l
                S[table-format=1.4]
                S[table-format=1.4]
                S[table-format=1.4]
                S[table-format=2.2]
            }
            \toprule
            {Plane} & {\(x\) (m)\tnote{a}} & {\(y\) (m)\tnote{b}} & {\(z\) (m)\tnote{c}} & {\(F\) (N)\tnote{d}} \\
            \midrule
            xy       & 0.0244 & 0.0236 & 0.0095 &  5.05 \\
            yz       & 0.0100 & 0.0181 & 0.0824 & 10.35 \\
            xz       & 0.0188 & 0.0093 & 0.0648 &  9.18 \\
            oblique  & 0.0109 & 0.0173 & 0.0689 &  9.15 \\
            \bottomrule
        \end{tabular*}
        \begin{tablenotes}
            \footnotesize
            \item[a] RMSE in \(x\)-direction.
            \item[b] RMSE in \(y\)-direction.
            \item[c] RMSE in \(z\)-direction.
            \item[d] RMSE of desired internal force.
        \end{tablenotes}
    \end{threeparttable}
\end{table}

\subsubsection{Octahedral Configuration}

With the octahedron configuration, we evaluate a configuration that could be used for both locomotion and manipulation making the task-space more interesting \cite{bae2023locomotion}.
Here, the VTT moves one box from one pile to another pile located \SI{0.5}{m} away, then places a second box on top of the first.
Both box piles are located inside the octahedral VTT structure when the manipulation begins.

The target objects are cube-shaped boxes, each weighing \SI{309}{g}. 
The grasping force is set to \SI{12}{N}, and the average manipulation speed is \SI{0.01}{m/s}. 
The manipulation sequence is as follows:
\begin{itemize}
    \item Move to the initial position using pure length control
    \item Grasp the first box with \SI{12}{N} grasping force
    \item Move in the $z$ direction by \SI{0.08}{m} for \SI{10}{s}
    \item Move in the $y$ direction by \SI{0.5}{m} for \SI{52}{s}
    \item Move in the $z$ direction by \SI{-0.4}{m} for \SI{42.5}{s}
    \item Release the first box
    \item Return to the second grasping position using pure position control
    \item Grasp the second box with \SI{12}{N} grasping force
    \item Move in the $z$ direction by \SI{0.1}{m} for \SI{8.5}{s}
    \item Move in the $y$ direction by \SI{0.5}{m} for \SI{52.5}{s}
    \item Release the second box
\end{itemize}

\begin{figure}[t]
\centering
\setlength{\tabcolsep}{0pt} 
\renewcommand{\arraystretch}{0} 

\newcommand{\Hgap}{0.02\textwidth}  
\newcommand{\Vgap}{4pt}             

\begin{tabular}{@{}c@{\hspace{\Hgap}}c@{}}

\begin{subfigure}[t]{0.35\textwidth}
  \includegraphics[width=\linewidth]{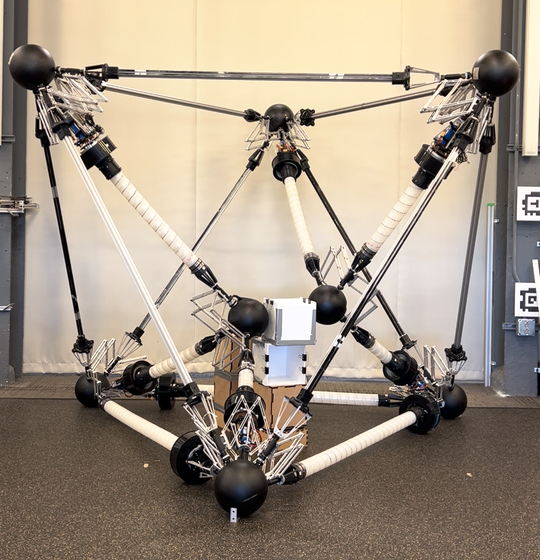}
  \caption{}
\end{subfigure}
&
\begin{subfigure}[t]{0.35\textwidth}
  \includegraphics[width=\linewidth]{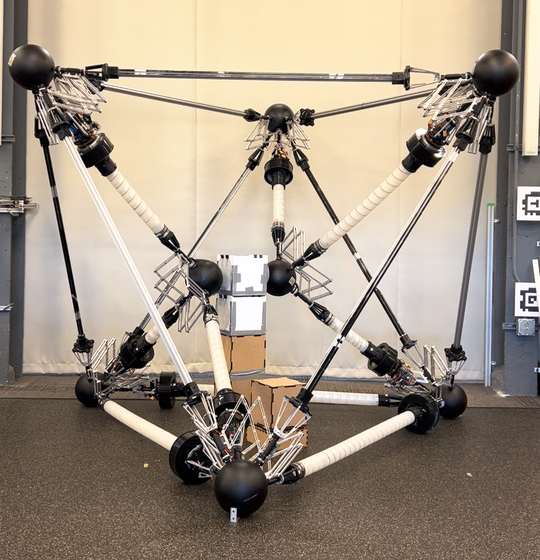}
  \caption{}
\end{subfigure}
\\[\Vgap]

\begin{subfigure}[t]{0.35\textwidth}
  \includegraphics[width=\linewidth]{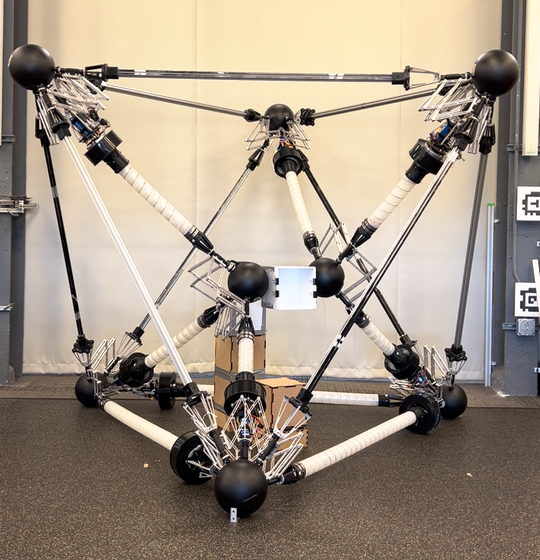}
  \caption{}
\end{subfigure}
&
\begin{subfigure}[t]{0.35\textwidth}
  \includegraphics[width=\linewidth]{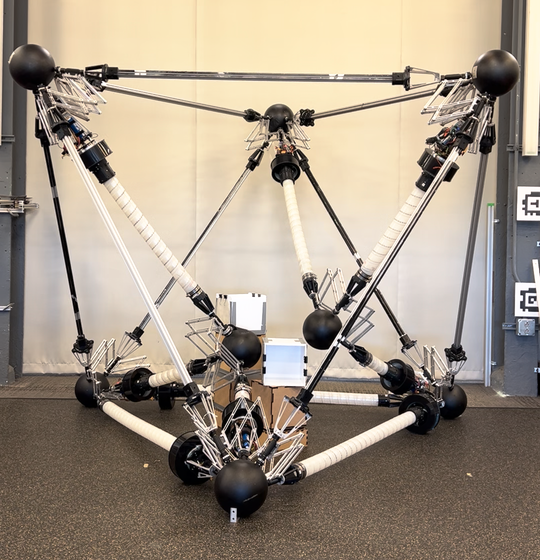}
  \caption{}
\end{subfigure}
\\[\Vgap]

\begin{subfigure}[t]{0.35\textwidth}
  \includegraphics[width=\linewidth]{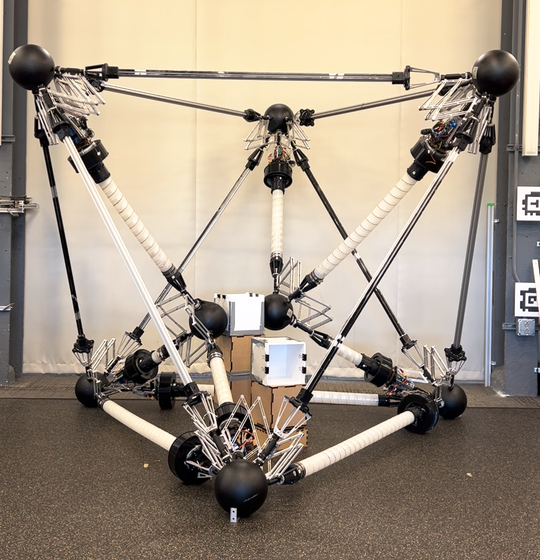}
  \caption{}
\end{subfigure}
&
\begin{subfigure}[t]{0.35\textwidth}
  \includegraphics[width=\linewidth]{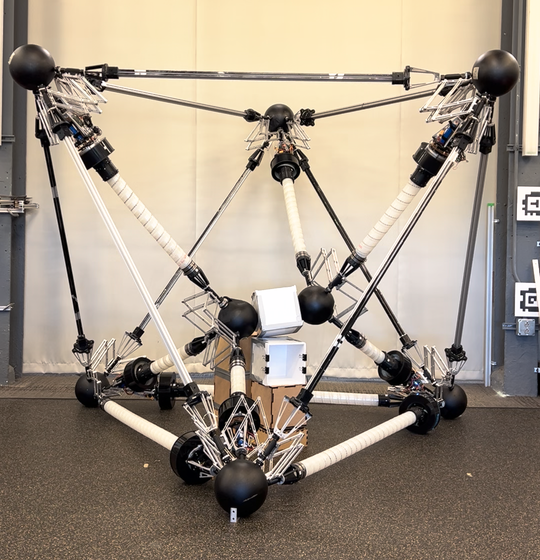}
  \caption{}
\end{subfigure}

\end{tabular}

\caption{Snapshots of the double tetrahedron VTT manipulation experiment. (a) 0:00~min; (b) 1:13~min; (c) 1:52~min; (d) 3:31~min; (e) 4:23~min; (f) 4:40~min.}
\label{Fig_Octahedron_Exp}
\end{figure}

Figure~\ref{Fig_Octahedron_Exp} shows snapshots of the two-box transfer experiment, and a supplementary video is provided.
The manipulation takes about 4~min~50~s to complete. 
This octahedron-based manipulation experiment demonstrates that VTT manipulation is repeatable, as the system successfully moves two objects consecutively.
With this demonstration, it is easy to see that the VTT can manipulate and carry objects through locomotion \cite{bae2023locomotion}.

\section{Conclusion}

In this paper, we proposed the concept of performing manipulation with the VTT system and demonstrated its feasibility. 
To enable manipulation, we developed a hierarchical hybrid controller capable of regulating position and force simultaneously without decoupling. 
On the hardware side, we improved the spiral zipper actuator design to achieve the required strength and tolerance. 
Because the spiral zipper actuator exhibits high friction and a large gear ratio, we implemented sensor-based force control for each truss member module. 
The force-tracking performance of the spiral zipper actuators was validated in multiple configurations. 
Manipulation performance was evaluated using the double-tetrahedron configuration, where we showed that the position tracking accuracy is sufficient for reliable manipulation while maintaining the grasping force. 
Finally, we demonstrated a practical manipulation scenario with the octahedron configuration.

\section{Funding Statement}
This research was supported by the MOTIE (Ministry of Trade, Industry, and Energy) in Korea, under the Human Resource Development Program for Industrial Innovation(Global) (P0017306, Global Human Resource Development for Innovative Design in Robot and Engineering) supervised by the Korea Institute for Advancement of Technology (KIAT).

\section{Acknowledgment}
Thanks to Henry Zhao for assisting the VTT programming and experiments.

\bibliographystyle{unsrt}  
\bibliography{references} 

\end{document}